\documentclass[conference,letter]{IEEEtran}
\IEEEoverridecommandlockouts

\usepackage[hidelinks]{hyperref}
\usepackage[cmex10]{amsmath}
\usepackage{amssymb,amsfonts}
\usepackage{dblfloatfix}

\usepackage[ruled,vlined]{algorithm2e}
\usepackage{graphicx}
\graphicspath{{Figures/PDF/}{Figures/PNG/}}

\usepackage{booktabs}
\usepackage{siunitx}
\usepackage[numbers,compress]{natbib}
\usepackage{texnames}
\usepackage{bm,bbm}
\usepackage{orcidlink}

\begin{document}

\title{
    \uppercase{Uncertainty Quantification in Surface Landmines and UXO Classification Using MC Dropout}

    \thanks{This work has been accepted and presented at IGARSS 2025 and will appear in the IEEE IGARSS 2025 proceedings.}

    \thanks{Part of this research has received support from the IEEE GRSS under the``ProjNET'' program.}
}





\author{ 	\IEEEauthorblockN{
                Sagar Lekhak \orcidlink{0009-0009-7896-6167}$^1$,
                Emmett J. Ientilucci \orcidlink{0000-0002-3643-8245}$^1$
                Dimah Dera \orcidlink{0000-0002-7168-5858}$^1$
                Susmita Ghosh\orcidlink{0000-0002-1691-761X}$^2$
            } 
            \\           
	      \IEEEauthorblockA{
                \textit{$^1$Rochester Institute of Technology,}
                Chester F. Carlson Center for Imaging Science,
		      Rochester, NY 14623, USA \\
                \textit{$^2$Jadavpur University,}
                Department of Computer Science and Engineering,
                Kolkata, West Bengal 700032, India
            }
}



\maketitle
\begin{abstract}


Detecting surface landmines and unexploded ordnances (UXOs) using deep learning has shown promise in humanitarian demining. However, deterministic neural networks can be vulnerable to noisy conditions and \emph{adversarial attacks}, leading to missed detection or misclassification. This study introduces the idea of uncertainty quantification through Monte Carlo (MC) Dropout, integrated into a fine-tuned ResNet-50 architecture for surface landmine and UXO classification, which was tested on a simulated dataset. Integrating the MC Dropout approach helps quantify epistemic uncertainty, providing an additional metric for prediction reliability, which could be helpful to make more informed decisions in demining operations. Experimental results on clean, adversarially perturbed, and noisy test images demonstrate the model's ability to flag unreliable predictions under challenging conditions. This proof-of-concept study highlights the need for uncertainty quantification in demining, raises awareness about the vulnerability of existing neural networks in demining to adversarial threats, and emphasizes the importance of developing more robust and reliable models for practical applications.


\end{abstract}


\section{Introduction}
Recent advancements in deep learning have shown promise for detecting surface landmines and UXOs using drone- or robot-based imagery. For instance, \citet{baur2021how_to_implement_drones}  applied a Faster R-CNN algorithm to detect PFM-1 mines from drone-acquired RGB imagery.
Similarly, \citet{deep_learning_based_yolov8} used YOLOv8-nano and YOLOv8-small to identify PFM-1 and PMA-2 mines in different environmental conditions.
Another study \cite{joint_fusion_and_detection_2023} combined YOLOv5 with image fusion techniques using RGB and near-infrared (NIR) multispectral data to detect mines with five different sizes and materials in various experimental scenes. 
While these studies focus on object detection, landmine classification remains a critical complementary task, especially when high-resolution imagery of isolated objects is available. In real-world scenarios, classification systems can aid in identifying the type of landmine (e.g., PFM-1, PMA-2, TM-62) once an object of interest has been localized, serving as a vital component in broader autonomous demining pipelines.

Despite advances in landmine detection, deep learning models remain vulnerable to adversarial examples: inputs subtly perturbed to mislead neural networks into incorrect predictions, such as misclassification or missed detections, while appearing unchanged to the human eye \cite{adversarial_attacks}. Physical adversarial perturbations, such as stickers on traffic signs \cite{eykholt2018robustphysicalworldattacksdeep} or patterns on clothing \cite{thys2019foolingautomatedsurveillancecameras}, have been shown to deceive models in real-world scenarios. In landmine detection and classification, similar perturbations on landmine surfaces could pose serious threats to model reliability. However, current approaches lack mechanisms to estimate model uncertainty under such adversarial or noisy conditions, revealing a significant gap in existing methods.

 

To address this gap, we present a study on applying Monte Carlo (MC) Dropout \cite{Gal2016Dropout} for uncertainty quantification in surface landmine and UXO classification—a domain where this approach has not been previously explored. By integrating the Bayesian theory of learning through MC Dropout into a classification pipeline, we aim to quantify the model’s predictive uncertainty and assess its potential for detecting adversarial attacks and perturbed inputs.

The main contributions of this work are as follows:





\begin{itemize}
    \item We adapt a ResNet-50 model \cite{resnet-50} pre-trained on ImageNet-1K \cite{imagenet-1k} and integrate MC Dropout, allowing predictive uncertainty estimation during inference.

    \item We evaluate the model across three scenarios of simulated data: (a) clean test images, (b) adversarial perturbations generated using the Fast Gradient Sign Method (FGSM) \cite{adversarial_attacks} and Projected Gradient Descent (PGD) \cite{madry2018towards}, and (c) noisy perturbations.

    \item We demonstrate that predictive uncertainty can serve as a valuable supplementary metric for surface landmine classification using a simulated dataset, facilitating more reliable decision-making under uncertain or adversarial conditions and laying the groundwork for future real-world demining applications.
\end{itemize}

To the best of our knowledge, this is the first application of uncertainty-aware, deep learning based classification for surface landmines and UXOs. This proof-of-concept aims to raise awareness of the need for uncertainty-aware approaches in surface landmine classification and detection, highlighting a critical gap in current deep learning methods.

\section{Methodology}

\subsection{Uncertainty Estimation using MC Dropout}
Deterministic neural networks do not inherently account for epistemic uncertainty, which arises due to limited data, lack of model knowledge, or non-optimal hyperparameter settings \cite{Gal2016Dropout}. Bayesian neural networks (BNNs) can estimate epistemic uncertainty in model weights by considering weights as \emph{probabilistic distributions} rather than deterministic values. However, BNNs are less practical for large-scale or high-dimensional models. BNNs are computationally expensive for larger models because they require approximating the \emph{posterior} distribution of weights, which involves integrating over high-dimensional parameter spaces. Exact inference is intractable, and approximation methods like Markov Chain Monte Carlo (MCMC) \cite{Robert2004MCMC} are slow and scale poorly, while Variational Inference (VI) \cite{Blei2017VI} simplifies the problem but struggles with complex \emph{posterior} distributions and requires significant computational resources.




Gal and Ghahramani [7] showed that enabling dropout during both training and testing can be approximated as variational inference in Bayesian Neural Networks (BNNs), and using dropout in this manner is referred to as \emph{Monte Carlo (MC) Dropout}. It provides a computationally efficient alternative to Bayesian inference. During inference, dropout is typically turned off, but \citet{Gal2016Dropout} proposed using dropout at test time as well. 
In this setting, they showed that the output becomes a stochastic function of the weights, and multiple stochastic forward passes approximate the \emph{posterior} distribution over the weights.
This approach involves running multiple stochastic forward passes through the network during inference, keeping dropout layers enabled. The average prediction over all forward passes gives the prediction of the model while the variance across these multiple forward stochastic passes provides an estimate of the epistemic uncertainty. If the variance across these predictions is high, it indicates uncertainty whereas low variance indicates high confidence in the prediction.



The authors model the process of dropping out each unit as a Bernoulli random variable \( z_i \), which is drawn from a Bernoulli distribution:

\begin{equation}
    z_i \sim \text{Bernoulli}(p),
\end{equation}

Here, \( p \) is the probability of retaining a neuron, and \( 1 - p \) is the probability of dropping the neuron. The network's output, for each forward pass with dropout, depends on which units are retained and which are dropped, and is given by:

\begin{equation}
    \hat{y}_\text{dropout}(x) = f(x, W, z),
\end{equation}
where \( f(x, W, z) \) is the neural network function with weights \( W \) and dropout mask \( z \).

By performing multiple forward passes with different dropout masks \( z \), we approximate the \emph{posterior} distribution over the weights. Specifically, the \emph{posterior} predictive distribution is approximated by:

\begin{equation}
    p(\hat{y} | x) \approx \frac{1}{N} \sum_{i=1}^{N} f(x, W, z_i),
\end{equation}
where \( N \) is the number of forward passes and \( f(x, W, z_i) \) is the output for the \( i \)-th forward pass with dropout mask \( z_i \).

The variance across these multiple stochastic forward passes provides an estimate of the epistemic uncertainty (\emph{i.e.,} model uncertainty). 

\begin{equation}
    \text{Variance}(\hat{y}) = \frac{1}{N} \sum_{i=1}^{N} \left(f(x, W, z_i) - \hat{y}_\text{avg}\right)^2,
\end{equation}
where \( \hat{y}_\text{avg} = \frac{1}{N} \sum_{i=1}^{N} f(x, W, z_i) \) is the average prediction over all forward passes which is the prediction of the model.  Here, the variance provides an estimate of the epistemic uncertainty associated with the prediction.

In our research, we utilized the high variance concept in MC Dropout to indicate high uncertainty in model predictions.


\subsection{Network Design and Optimization}

We utilized a ResNet-50 model pretrained on ImageNet-1K from the PyTorch framework, consisting of a stem layer and four residual layer groups, denoted as, layer1 to layer4 in PyTorch and referred to as Block 1 to Block 4 in this paper. These blocks extract progressively higher-level features, with Block 1 capturing low-level features and Block 4 capturing high-level semantic features. We modified the final Fully Connected (FC) layer of ResNet-50 for our four classes, experimented with adding intermediate FC layers, multiple dropout layers, various dropout rates, learning rates, and configurations of frozen and unfrozen blocks. 

Despite these efforts, the training and validation accuracies were initially low, with unstable learning curves observed. Higher accuracies were achieved with a single dropout layer rather than multiple dropout layers with high dropout rates. Only a subset of the configurations are listed in  Table \ref{tab:experiment_results} due to manuscript space constraints.

\begin{table}[hbt]
   \centering
   \caption{Training and validation results for different dropout rates and layer configurations.}
   \label{tab:experiment_results}
   \small 
   \setlength{\tabcolsep}{4pt} 
   \begin{tabular}{@{}l S[table-format=1.1] S[table-format=1.2] S[table-format=1.2] S[table-format=1.2] S[table-format=1.2]@{}}
       \toprule
       \shortstack{\textbf{ResNet-50} \\ \textbf{Blocks} \\ \textbf{Unfrozen}} & \shortstack{\textbf{Drop-}\\ \textbf{out} \\ \textbf{Rate}} & \textbf{Train Acc.} & \textbf{Train Loss} & \textbf{Val Acc.} & \textbf{Val Loss} \\ 
       \cmidrule(lr){1-1} \cmidrule(lr){2-2} \cmidrule(lr){3-3} \cmidrule(lr){4-4} \cmidrule(lr){5-5} \cmidrule(lr){6-6}
       None         & 0.1 & 0.75 & 0.61 & 0.88 & 0.33 \\
       None         & 0.6 & 0.94 & 0.12 & 0.88 & 0.86 \\
       Only 4       & 0.1 & 0.94 & 0.13 & 0.99 & 0.01 \\
       3 and 4      & 0.0 & 0.94 & 0.11 & 0.94 & 0.26 \\
       \textbf{3 and 4}         & \textbf{0.1} & \textbf{0.95} & \textbf{0.12} & \textbf{0.99} & \textbf{0.02} \\
       3 and 4      & 0.2 & 0.96 & 0.10 & 0.99 & 0.01 \\
       3 and 4      & 0.3 & 0.95 & 0.12 & 0.99 & 0.01 \\
       3 and 4      & 0.6 & 0.95 & 0.12 & 0.99 & 0.03 \\
       \bottomrule
   \end{tabular}
\end{table}

As seen in the first row of  Table \ref{tab:experiment_results}, when no layers were unfrozen, the training and validation accuracies dropped significantly. In the second row, with a high dropout rate, the validation accuracy was lower than the training accuracy, and the validation loss was much higher, indicating overfitting. Although similar accuracies were achieved with different dropout rates after unfreezing the blocks three and four of ResNet-50, the bolded configuration shown in the table was selected because it demonstrated stable learning curves and better generalization compared to other configurations.

\begin{figure}[hbt]
	\centering
	\includegraphics[width=.9\linewidth]{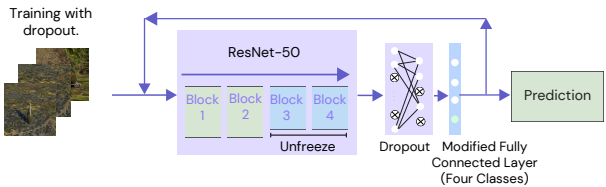}
	\caption{Proposed modified ResNet-50 network architecture during training phase with last two layers unfrozen, modified final Fully Connected (FC) layer and a dropout layer before the final FC layer.}\label{fig:training_pdf}
\end{figure}

The final modified ResNet-50 architecture, during training, is shown in Fig. \ref{fig:training_pdf}. 
The model was trained on an NVIDIA RTX A4000 GPU using 5,208 training images for 100 epochs, a batch size of 16, a learning rate of 0.0001, and a dropout rate of 0.1. The total training time was 25 minutes and 30 seconds.

\begin{figure}[hbt]
	\centering
	\includegraphics[width=.9\linewidth]{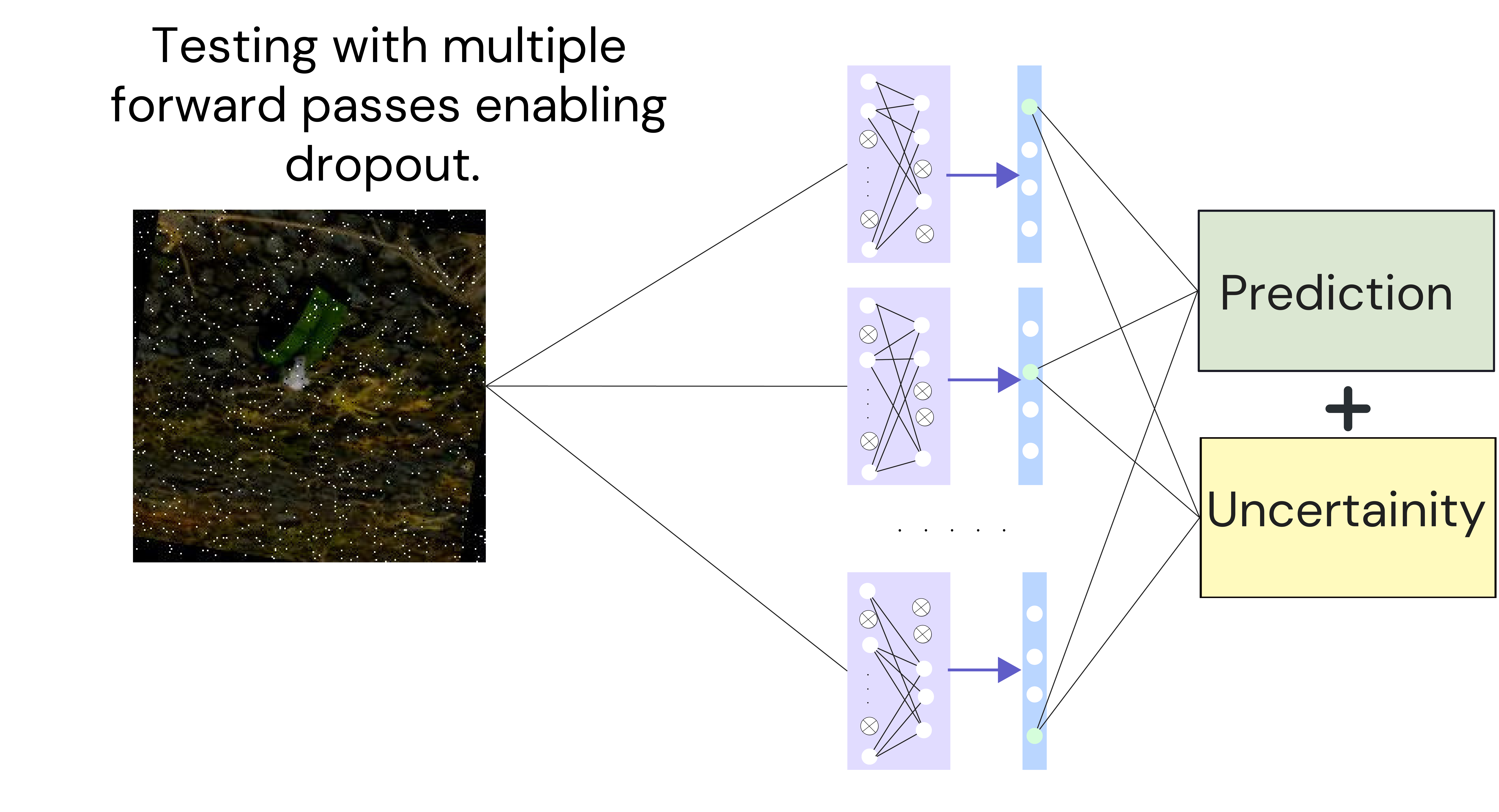}
	\caption{Illustration of the testing phase, where multiple stochastic forward passes are performed on a trained model with the dropout layer enabled, enabling uncertainty estimation for the test sample.}\label{fig:testing_pdf}
\end{figure}

During inference, Monte Carlo (MC) Dropout was applied by retaining the dropout layer and performing 100 stochastic forward passes. This allowed the model to generate a distribution of outputs for each test sample, capturing uncertainty. The final prediction was obtained by computing the mean of these outputs, with the variance serving as a measure of uncertainty, which indicates the confidence (higher uncertainty, less confidence) in the model's predictions. A visual representation of this process is shown in Fig. \ref{fig:testing_pdf}.

\section{Results}
In this section, we present the results of our experiments to evaluate the effectiveness of the proposed MC Dropout-based uncertainty quantification approach for landmine classification. The evaluation is carried out across three scenarios: (1) clean test images, (2) adversarially perturbed inputs, and (3) noisy samples. We first describe the dataset used in our experiments and then discuss the results for each scenario in detail.
\subsection{Dataset}

Landmine datasets are often not made publicly available due to security and defense concerns. To demonstrate our proof-of-concept, we utilized a publicly available simulated dataset from the Roboflow Universe platform \cite{dataset}. The dataset contains 5,952 images split into training (5,208), validation (496), and test (248) sets, with four object classes: grenade (class 0), landmine (class 1), projectile (class 2), and rocket (class 3). Class distribution is relatively balanced across all splits, with each class representing approximately 23–28\% of the total images, ensuring consistent class representation for training and evaluation. The augmentation and preprocessing techniques applied to the training set are listed in \cite{dataset}; however, the specific type of noise used is not clearly documented. Upon careful inspection, it was observed that noise was added exclusively to the training data, while the validation and test sets remain noise-free.

\subsection{Predictive Uncertainty on Clean Test Images}
\begin{figure}[hbt]
	\centering
	\includegraphics[width=.9\linewidth]{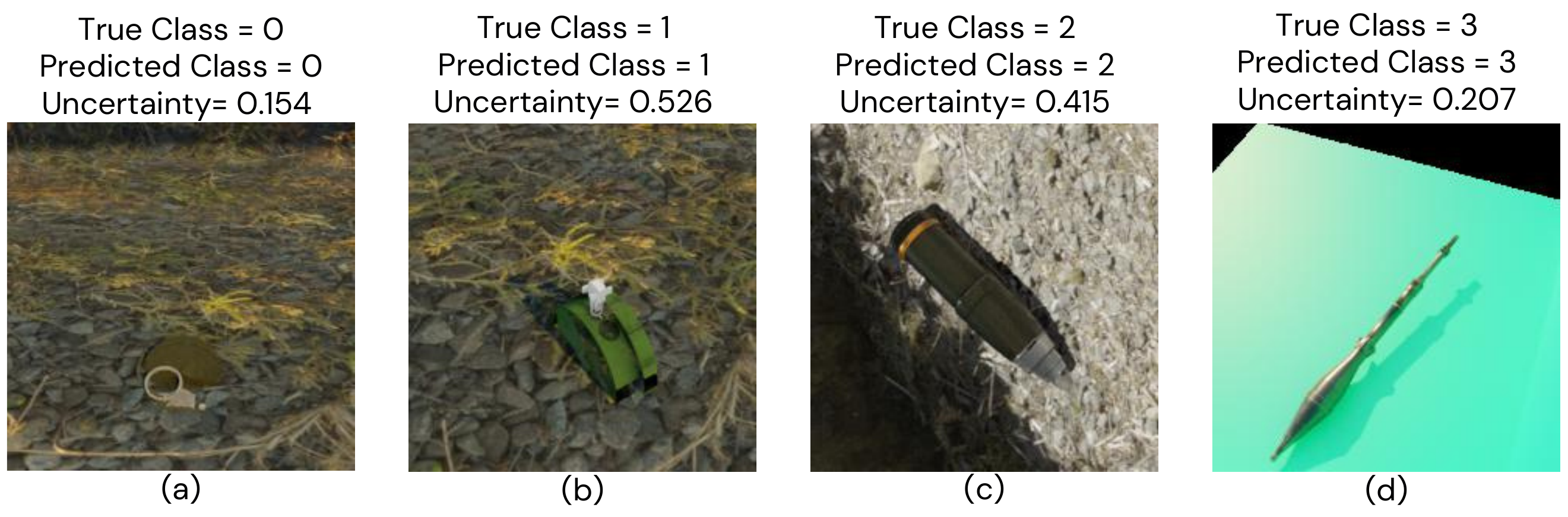}
	\caption{Predictive uncertainty on randomly selected clean test samples, with low uncertainty (\emph{i.e.,} high confidence) in (a) and (d), and higher uncertainty in (b) and (c).}\label{fig:result_1}
\end{figure}

For the clean test samples, the model correctly classified all inputs but exhibited varying levels of uncertainty. Four randomly selected samples from each class in the test set are shown in Fig. \ref{fig:result_1}, where the model correctly predicted all four samples. For samples (a) and (d), the model exhibited high confidence (\emph{i.e.,} low uncertainty), while for samples (b) and (c), the model displayed higher uncertainty in its predictions. This behavior highlights the model’s capability to quantify uncertainty even when predictions are correct, particularly for more challenging inputs.

\subsection{Predictive Uncertainty on Adversarially Perturbed Inputs}

\begin{figure}[hbt]
	\centering
	\includegraphics[width=.9\linewidth]{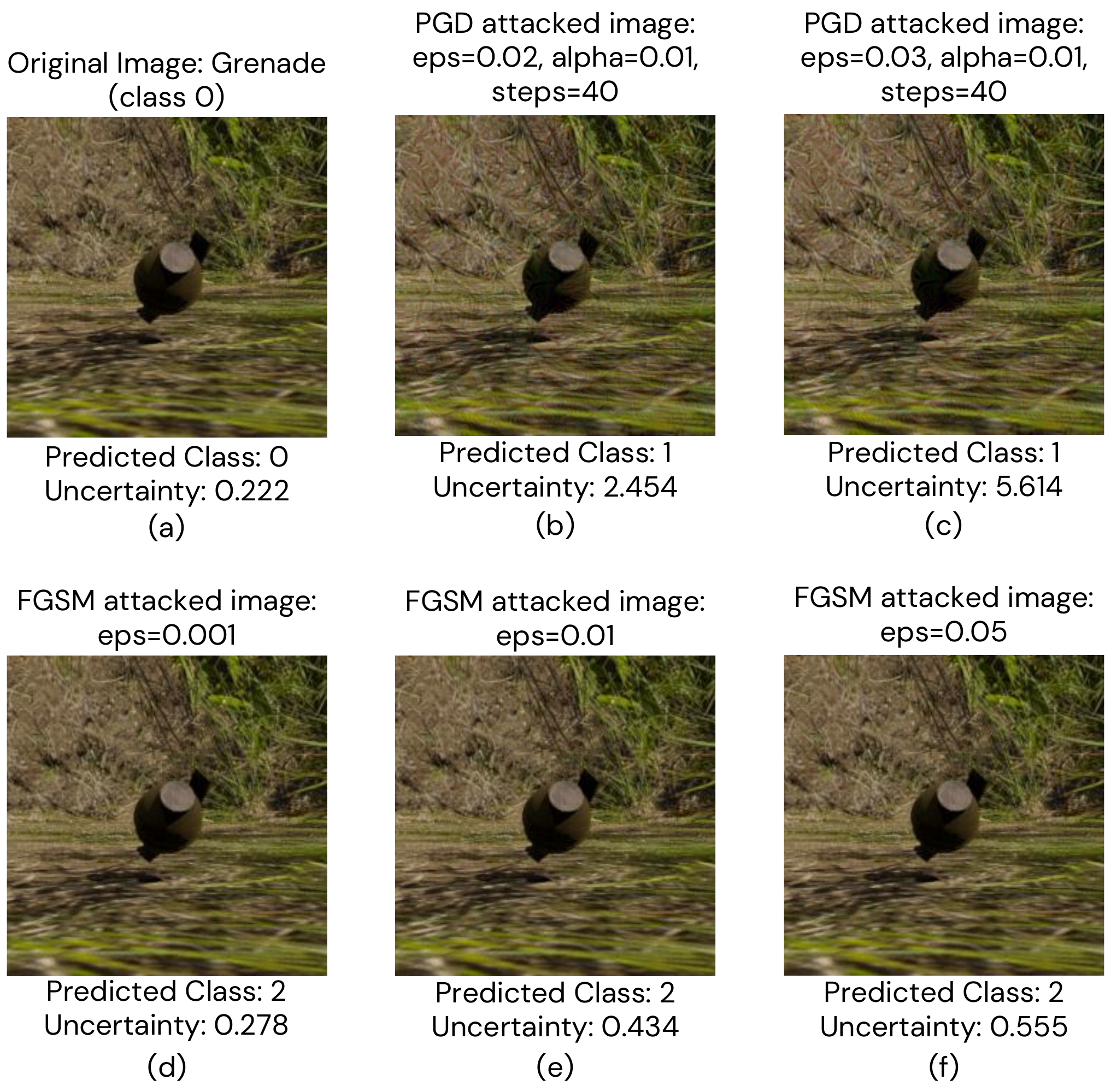}
	\caption{Model predictions and uncertainties with varying level of adversarial perturbations. (a) Clean image. (b) PGD attack ($\epsilon = 0.02$). (c) PGD attack ($\epsilon = 0.03$). (d) FGSM attack ($\epsilon = 0.001$) (e) FGSM attack ($\epsilon = 0.01$). (f) FGSM attack ($\epsilon = 0.05$). }\label{fig:result_2}
\end{figure}

To evaluate model robustness against adversarial attacks, we tested it on inputs perturbed by PGD \cite{madry2018towards} and FGSM \cite{adversarial_attacks} attacks. A clean image from class 0 was correctly classified with low uncertainty 0.222, as shown in Fig. \ref{fig:result_2}(a). Although all the images in Figs. \ref{fig:result_2}(b) -  \ref{fig:result_2}(f) appear similar to human eye, they were perturbed and easily fooled the model. Adversarial perturbations caused misclassification of the true class but resulted in significantly higher uncertainty values, indicating unreliable predictions.  Under PGD attacks, as shown in Figs. \ref{fig:result_2}(b) and \ref{fig:result_2}(c), the model misclassified the original image as class 1, with the uncertainty increasing to 2.454 for $\epsilon = 0.02$ and 5.614 for $\epsilon = 0.03$. FGSM attacks showed a similar trend, as shown in Figs. \ref{fig:result_2}(d) - \ref{fig:result_2}(f), with the same image misclassified as class 2. The uncertainty rose from 0.278 at $\epsilon = 0.001$ to 0.434 for $\epsilon = 0.01$ and 0.555 for $\epsilon = 0.05$.

These results presented here demonstrate the model signals unreliable predictions under adversarial perturbations, with PGD attacks exhibiting a more pronounced effect.

\subsection{Predictive Uncertainty on Noisy Samples}

\begin{figure}[hbt]
	\centering
	\includegraphics[width=.9\linewidth]{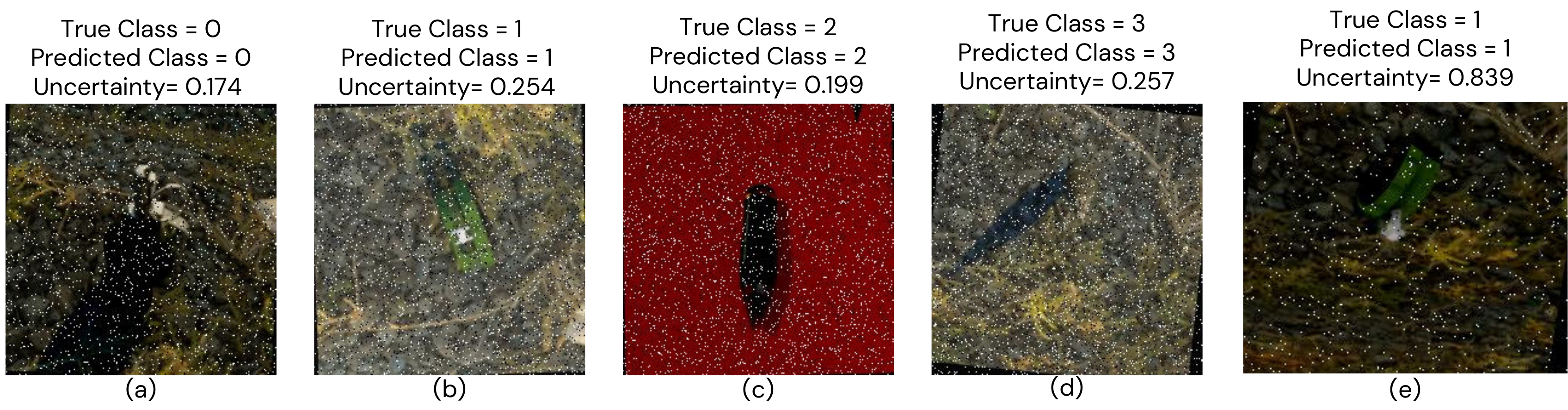}
	\caption{Predictive uncertainty on randomly selected noisy training samples, with low uncertainty in (a), (b), (c) and (d), and higher uncertainty in (e).}\label{fig:results_3}
\end{figure}

Some noisy samples were randomly selected from each class in the training set, as noise was only introduced during data augmentation in the training split, as provided in the dataset \cite{dataset}. Interestingly, some correctly classified but noisy training samples still exhibited higher predictive uncertainty, highlighting the model’s ability to recognize ambiguity despite having seen those examples during training. In Fig. \ref{fig:results_3}, the model correctly classified the true classes with low uncertainty for Figs. \ref{fig:results_3}(a) - \ref{fig:results_3}(d) while it exhibited higher uncertainty (\emph{i.e.,} 0.839) in Fig. \ref{fig:results_3}(e). 


\section{Discussion}

This proof-of-concept study explores the feasibility and potential value of incorporating uncertainty quantification, via Monte Carlo Dropout, into surface landmine and UXO classification. The findings suggest that as adversarial noise in the data increases, the uncertainty in predictions, or variance, tends to increase as well, which could help support more informed decision-making in humanitarian demining efforts. Existing deep learning methods for surface landmine detection are susceptible to both false negatives (\emph{i.e.,} missed detections or type II errors) and false positives (\emph{i.e.,} false alarms or type I errors). While false negatives pose significant safety risks, false positives, though less critical to safety, can reduce operational efficiency by causing delays and increasing costs. Our work aims to raise awareness within the demining community about these challenges and emphasizes the potential role of uncertainty estimation in helping to mitigate these risks in real-world demining scenarios, where adverse conditions and uncertainty are common. 

While the results are promising, we acknowledge that the choice of dropout rate (0.1 in this study) and fine-tuning strategy were based on limited experimentation. Since MC Dropout introduces stochasticity during inference, the predictive distributions — and thus uncertainty estimates — can vary across runs and dropout configurations. Parameter optimization (e.g., dropout rate tuning, learning rate, and layer unfreezing strategies) remains an important direction for future work, especially when transitioning from simulated to real-world datasets, where robust uncertainty estimation will be even more critical for practical deployment.
We also recognize that our approach may not be universally effective against all types of adversarial attacks, and it might not produce the same trend of increased variance (uncertainty) with different levels of noise introduced by various adversarial perturbations. Therefore, studying how variance trends with respect to different adversarial noise levels is an equally important aspect to consider. We believe it is more meaningful to conduct such tuning and report associated uncertainty metrics on real-world datasets, where environmental variability and sensor noise more closely resemble operational scenarios.  

Further validation of the proposed uncertainty quantification approach using real-world datasets could support the establishment of uncertainty thresholds to flag unreliable predictions, helping to better understand both its potential and limitations. Due to space constraints, this paper presents an initial framework, while ongoing work focuses on more robust analysis using real-world data. Future research could extend this and other uncertainty estimation approaches to state-of-the-art object detection models—such as DETR, YOLO, and Faster R-CNN—applied to real-world surface landmine datasets.

\small
\bibliographystyle{IEEEtranN}
\bibliography{references}

\begin{thebibliography}{13}
\providecommand{\natexlab}[1]{#1}
\providecommand{\url}[1]{#1}
\csname url@samestyle\endcsname
\providecommand{\newblock}{\relax}
\providecommand{\bibinfo}[2]{#2}
\providecommand{\BIBentrySTDinterwordspacing}{\spaceskip=0pt\relax}
\providecommand{\BIBentryALTinterwordstretchfactor}{4}
\providecommand{\BIBentryALTinterwordspacing}{\spaceskip=\fontdimen2\font plus
\BIBentryALTinterwordstretchfactor\fontdimen3\font minus \fontdimen4\font\relax}
\providecommand{\BIBforeignlanguage}[2]{{%
\expandafter\ifx\csname l@#1\endcsname\relax
\typeout{** WARNING: IEEEtranN.bst: No hyphenation pattern has been}%
\typeout{** loaded for the language `#1'. Using the pattern for}%
\typeout{** the default language instead.}%
\else
\language=\csname l@#1\endcsname
\fi
#2}}
\providecommand{\BIBdecl}{\relax}
\BIBdecl

\bibitem[Baur et~al.(2021)Baur, Steinberg, Nikulin, Chiu, and de~Smet]{baur2021how_to_implement_drones}
\BIBentryALTinterwordspacing
J.~Baur, G.~Steinberg, A.~Nikulin, K.~Chiu, and T.~de~Smet, ``How to implement drones and machine learning to reduce time, costs, and dangers associated with landmine detection,'' \emph{The Journal of Conventional Weapons Destruction}, vol.~25, no.~1, 2021. [Online]. Available: \url{https://commons.lib.jmu.edu/cisr-journal/vol25/iss1/29}
\BIBentrySTDinterwordspacing

\bibitem[Vivoli et~al.(2024)Vivoli, Bertini, and Capineri]{deep_learning_based_yolov8}
\BIBentryALTinterwordspacing
E.~Vivoli, M.~Bertini, and L.~Capineri, ``Deep learning-based real-time detection of surface landmines using optical imaging,'' \emph{Remote Sensing}, vol.~16, no.~4, 2024. [Online]. Available: \url{https://www.mdpi.com/2072-4292/16/4/677}
\BIBentrySTDinterwordspacing

\bibitem[Qiu et~al.(2023)Qiu, Guo, Hu, Jiang, and Luo]{joint_fusion_and_detection_2023}
Z.~Qiu, H.~Guo, J.~Hu, H.~Jiang, and C.~Luo, ``Joint fusion and detection via deep learning in uav-borne multispectral sensing of scatterable landmine,'' \emph{Sensors (Basel)}, vol.~23, no.~12, p. 5693, 2023.

\bibitem[Goodfellow et~al.(2015)Goodfellow, Shlens, and Szegedy]{adversarial_attacks}
\BIBentryALTinterwordspacing
I.~J. Goodfellow, J.~Shlens, and C.~Szegedy, ``Explaining and harnessing adversarial examples,'' in \emph{3rd International Conference on Learning Representations, {ICLR} 2015, San Diego, CA, USA, May 7-9, 2015, Conference Track Proceedings}, Y.~Bengio and Y.~LeCun, Eds., 2015. [Online]. Available: \url{http://arxiv.org/abs/1412.6572}
\BIBentrySTDinterwordspacing

\bibitem[Eykholt et~al.(2018)Eykholt, Evtimov, Fernandes, Li, Rahmati, Xiao, Prakash, Kohno, and Song]{eykholt2018robustphysicalworldattacksdeep}
K.~Eykholt, I.~Evtimov, E.~Fernandes, B.~Li, A.~Rahmati, C.~Xiao, A.~Prakash, T.~Kohno, and D.~Song, ``Robust physical-world attacks on deep learning visual classification,'' in \emph{Proceedings of the IEEE conference on computer vision and pattern recognition}, 2018, pp. 1625--1634.

\bibitem[Thys et~al.(2019)Thys, Ranst, and Goedemé]{thys2019foolingautomatedsurveillancecameras}
S.~Thys, W.~V. Ranst, and T.~Goedemé, ``Fooling automated surveillance cameras: Adversarial patches to attack person detection,'' in \emph{2019 IEEE/CVF Conference on Computer Vision and Pattern Recognition Workshops (CVPRW)}, 2019, pp. 49--55.

\bibitem[Gal and Ghahramani(2016)]{Gal2016Dropout}
Y.~Gal and Z.~Ghahramani, ``Dropout as a bayesian approximation: Representing model uncertainty in deep learning,'' in \emph{international conference on machine learning}.\hskip 1em plus 0.5em minus 0.4em\relax PMLR, 2016, pp. 1050--1059.

\bibitem[He et~al.(2016)He, Zhang, Ren, and Sun]{resnet-50}
K.~He, X.~Zhang, S.~Ren, and J.~Sun, ``Deep residual learning for image recognition,'' in \emph{2016 IEEE Conference on Computer Vision and Pattern Recognition (CVPR)}, 2016, pp. 770--778.

\bibitem[Deng et~al.(2009)Deng, Dong, Socher, Li, Li, and Fei-Fei]{imagenet-1k}
J.~Deng, W.~Dong, R.~Socher, L.-J. Li, K.~Li, and L.~Fei-Fei, ``Imagenet: A large-scale hierarchical image database,'' in \emph{2009 IEEE Conference on Computer Vision and Pattern Recognition}, 2009, pp. 248--255.

\bibitem[Madry et~al.(2018)Madry, Makelov, Schmidt, Tsipras, and Vladu]{madry2018towards}
\BIBentryALTinterwordspacing
A.~Madry, A.~Makelov, L.~Schmidt, D.~Tsipras, and A.~Vladu, ``Towards deep learning models resistant to adversarial attacks,'' in \emph{International Conference on Learning Representations}, 2018. [Online]. Available: \url{https://openreview.net/forum?id=rJzIBfZAb}
\BIBentrySTDinterwordspacing

\bibitem[Robert and Casella(2004)]{Robert2004MCMC}
C.~P. Robert and G.~Casella, \emph{Monte Carlo Statistical Methods}, 2nd~ed.\hskip 1em plus 0.5em minus 0.4em\relax Springer, 2004.

\bibitem[Blei et~al.(2017)Blei, Kucukelbir, and McAuliffe]{Blei2017VI}
\BIBentryALTinterwordspacing
D.~M. Blei, A.~Kucukelbir, and J.~D. McAuliffe, ``Variational inference: A review for statisticians,'' \emph{Journal of the American Statistical Association}, vol. 112, no. 518, pp. 859--877, 2017. [Online]. Available: \url{https://doi.org/10.1080/01621459.2017.1285773}
\BIBentrySTDinterwordspacing

\bibitem[Ransome()]{dataset}
\BIBentryALTinterwordspacing
T.~Ransome, ``Ordnance id dataset.'' [Online]. Available: \url{https://universe.roboflow.com/taamir-ransome/ordnance-id}
\BIBentrySTDinterwordspacing

\end{thebibliography}

\end{document}